\documentclass{ieeetran}
\usepackage{cite}
\usepackage{amsmath,amssymb,amsfonts}
\usepackage{algorithmic}
\usepackage{graphicx}
\usepackage{textcomp}
\usepackage{bm}
\usepackage{url}
\usepackage{threeparttable}
\usepackage{tabularx}
\usepackage{hyperref}
\usepackage{verbatim} 
\usepackage{tikz}
\usepackage{fancyvrb}
\hypersetup{
  colorlinks = true, 
  linkcolor = blue,  
  citecolor = blue, 
  filecolor = magenta, 
  urlcolor = cyan    
}
\usetikzlibrary{arrows.meta, positioning}
\def\BibTeX{{\rm B\kern-.05em{\sc i\kern-.025em b}\kern-.08em
    T\kern-.1667em\lower.7ex\hbox{E}\kern-.125emX}}

\title{Efficient Standardization of Clinical Notes using Large Language Models}

\author{
    Daniel B. Hier, \IEEEmembership{Senior Member, IEEE}, 
    Michael D. Carrithers, Thanh Son Do, 
    Tayo Obafemi-Ajayi, \IEEEmembership{Member, IEEE}
    \thanks{Daniel B. Hier is with the Kummer Institute Center for Artificial Intelligence and Autonomous Systems, Missouri University of Science \& Technology. (e-mail: hierd@umsystem.edu).}
    \thanks{Michael D. Carrithers is with the Department of Neurology and Rehabilitation, University of Illinois at Chicago, Chicago, IL 60612. (e-mail: mcar1@uic.edu).}
    \thanks{Thanh Son Do is with the Cooperative Engineering Program, Missouri State University, Springfield, MO. (e-mail: td64s@missouristate.edu).}
    \thanks{Tayo Obafemi-Ajayi is with the Cooperative Engineering Program, Missouri State University, Springfield, MO. (e-mail: tayoobafemiajayi@missouristate.edu).}
}

\begin{document}
\maketitle

\begin{abstract}
Clinician notes are a rich source of patient information but often contain inconsistencies due to varied writing styles, colloquialisms, abbreviations, medical jargon, grammatical errors, and non-standard formatting. These inconsistencies hinder the extraction of meaningful data from electronic health records (EHRs), posing challenges for quality improvement, population health, precision medicine, decision support, and research. 

We present a large language model approach to standardizing a corpus of 1,618 clinical notes. Standardization corrected an average of $4.9 \pm 1.8$ grammatical errors, $3.3 \pm 5.2$ spelling errors, converted $3.1 \pm 3.0$ non-standard terms to standard terminology, and expanded $15.8 \pm 9.1$ abbreviations and acronyms per note. Additionally, notes were re-organized into canonical sections with standardized headings. This process prepared notes for key concept extraction, mapping to medical ontologies, and conversion to interoperable data formats such as FHIR. 

Expert review of randomly sampled notes found no significant data loss after standardization. This proof-of-concept study demonstrates that standardization of clinical notes can improve their readability, consistency, and usability, while also facilitating their conversion into interoperable data formats.

\textit{Clinical relevance}—Standardizing clinician notes using large language models improves the quality, accessibility, and usability of critical patient information, supporting better patient outcomes and enhancing data interoperability.
\end{abstract}


\section{Introduction}
\label{sec:introduction}
\IEEEPARstart{E}lectronic Health Records (EHRs) have revolutionized healthcare documentation, improving both the legibility and accessibility of patient data \cite{menachemi2011benefits}. However, there has been a profound rise in physician dissatisfaction with EHRs and an associated increase in clinician burnout. EHRs have addressed two major challenges: the ``availability problem" (paper patient charts were frequently unavailable) and the ``legibility problem" (handwritten notes were notoriously difficult to read) \cite{bruner2001handwriting, kozak1994deciphering, rodriguez2002illegible}. Physician dissatisfaction stems from an onerous documentation burden, the shift of clerical tasks to physicians, and poor EHR design. Additional factors include complex EHR interfaces, excessive workloads, time pressures, fatigue from automated alerts, and the perception that EHRs do not enhance patient care quality. Many physicians also feel inadequately trained and supported in the use of their EHR \cite{muhiyaddin2022electronic, downing2018physician, elliott2022electronic, carroll2015health, rodriguez2022s, koopman2015physician, budd2023burnout}.

The use of large language models in healthcare is rapidly expanding. These models support critical applications such as decision support, diagnosis explanation, clinical text summarization, concept and information extraction, machine-coding of documents, concept mapping to ontologies, and the conversion of documents into exchangeable formats \cite{li2023enhancing, van2023clinical, tang2023evaluating, zhou2023considerations, qiu2023large, wang2023large}. By automating labor-intensive documentation processes, large language models help address the documentation burden.

We use ``standardization" to refer to the process of enhancing the structural and linguistic integrity of a clinical note without altering its clinical content. Standardization focuses on improving format, spelling, grammar, terminology, usability, interoperability, and readability. It includes replacing abbreviations (e.g., \texttt{OU} $\rightarrow$ \texttt{both eyes}), substituting conventional terms for colloquial terms (e.g., \texttt{upgoing toe} $\rightarrow$ \texttt{Babinski sign}), and correcting misspellings (e.g., \texttt{vscalar} $\rightarrow$ \texttt{vascular}). We hypothesize that a large language model such as GPT-4 can perform note standardization efficiently and accurately.

Physicians and other clinicians create clinical notes through direct text entry (typing), dictation, copy-and-paste from other notes, and automated text insertion from structured information held elsewhere in the EHR. Multiple problems plague the current use of clinical notes in the EHR \cite{hultman2019challenges}, including non-grammatical forms, colloquialisms, acronyms \cite{kugic2024disambiguation}, non-standard terminology, slang \cite{aronson2023use}, jargon \cite{castro2007babel, lee2014literature}, euphemisms, misspellings \cite{workman2019efficient}, and ambiguous language \cite{prater2019electronic, aronson2023use, castro2007babel, pitt2020eradicating, workman2019efficient}. Studies have shown that as many as 20\% of text is abbreviations \cite{hamiel2018frequency}, which range from unapproved and non-standard to cryptic and dangerous (e.g., MS for ``mental status", ``morphine sulfate," or ``multiple sclerosis") \cite{myers2011randomized, horon2012prohibited, cheung2018audit, shultz2011avoiding, baker2006campaign, american2005medication}. Notes may be poorly organized and lack proper headers for important sections such as the history, examination, impression, and plan \cite{koopman2015physician, hultman2019challenges}. Note-writing skills vary by clinician, with some adept while others struggle \cite{banda2018advances}. Most physician notes are free text and lack machine-codes from standard medical terminologies like ICD, LOINC, RxNorm, or SNOMED CT \cite{mcdonald2003loinc, hanna2013building, zarei2023comparison, lee2014literature}. Some notes are verbose, while others lack sufficient detail.

The standardization of clinical notes using large language models offers a path to address several documentation challenges that undermine the usability, quality, and exchangeability of physician notes. Key problems addressable by large language models include:
\begin{itemize}
    \item \textit{Expand abbreviations} to their full forms, mitigating confusion caused by ambiguous or non-standard acronyms (e.g., ``MS" expanded to ``multiple sclerosis" or ``mental status" based on context). Although ``alerts" have been tried to reduce the use of unapproved abbreviations, they have not been well-accepted and have not demonstrated changes in physician behavior \cite{myers2011randomized}.
    \item \textit{Correct typographical, spelling, and grammatical errors} to enhance note readability, quality, and professionalism \cite{ficarra1981grammar}.
    \item \textit{Identify and replace slang, colloquialisms, and other non-standard terms}, improving the precision and clarity of notes (e.g., ``heart attack" replaced by ``myocardial infarction"). Dictation, though widely used, often introduces unwanted word substitutions that change the meaning of text \cite{goss2016incidence}. Large language models can catch and correct these substitutions.
    \item \textit{Improve note organization} by structuring notes into canonical sections (e.g., HISTORY, EXAMINATION, IMPRESSION, PLAN), making them more navigable and interpretable for healthcare providers.
    \item \textit{Identify mappable medical entities} such as signs, symptoms, medications, and diagnoses that can be linked to machine codes from appropriate medical ontologies.
    \item \textit{Create outputs suitable for searches and data exchanges} such as JSON, enabling semi-structured data retrieval, integration into databases, and preparing notes for conversion into formats compatible with exchange standards such as HL7-FHIR \cite{li2023enhancing, bender2013hl7, vorisek2022fast}.
\end{itemize}

The goal of note standardization is to improve the structure, terminology, and readability of clinical notes without altering their underlying clinical content.

\begin{table}[!ht]
    \centering
    \caption{Quality Ratings for Key Metrics for standardized Notes}
    \begin{threeparttable}
    \begin{tabular}{lrr}
    \hline
        \textbf{Metric} & \textbf{Mean} & \textbf{SD} \\ \hline
        Text Organization & 4.93 & 0.43 \\ 
        Spelling and Grammar & 4.96 & 0.39 \\ 
        Abbreviation Expansion & 4.74 & 0.56 \\ 
        Terminology Standardization & 4.81 & 0.52 \\ 
        Completeness & 4.04 & 0.53 \\ \hline
    \end{tabular}
    \begin{tablenotes}
        \item[1] Quality ratings are based on a 5-point Likert scale, where 5 represents the highest quality.
        \item[2] Metrics were a consensus of a human expert and GPT-4.
        \item[3] SD = Standard Deviation.
    \end{tablenotes}
    \end{threeparttable}
    \label{normalization_metrics}
\end{table}

\section{Methods}
This study explores the application of GPT-4 to standardize 1618 de-identified clinical notes from a Neurology Clinic. We hypothesized that standardization can effectively improve the structure, clarity, and usability of physician notes while supporting downstream integration into alternative data formats. Our aim was to standardize a corpus of physician notes containing grammatical errors, spelling errors, excessive abbreviations, and poor formatting. The standardization process was executed using the GPT-4 API, guided by appropriate instructions.

\textit{Data Acquisition}:~~Consent was obtained from the University of Illinois IRB to use clinical notes from the Neurology Clinic for research purposes. All notes were deidentified by REDCap \cite{harris2009research}. Notes were selected to carry neuroimmunological diagnoses such as multiple sclerosis or Guillain-Barré syndrome. REDCap provided 21,028 notes for the years 2016 to 2022 with neuroimmunological diagnoses. We filtered those notes with the following requirements: outpatient notes only, notes from the Neurology Clinic only, physician notes only, and a length of at least 2000 characters. This yielded a final dataset of 1618 physician clinical notes (mean characters $6423 \pm 3689$). All were typed by the resident or attending physician; none were dictated or transcribed. REDCap provided us with a CSV file with a field for the note text as unformatted ASCII text. Each note was assigned an accession number and was saved to file as a JSON-compatible object with the format:
\begin{Verbatim}[fontsize=\footnotesize]
{
    "accession_num": "1",
    "note_text": 
    "NEUROLOGY CLINIC NOTE
    Chief Complaint: New onset of double vision.
    History: History of optic neuritis and numbness.
    Examination: Increased reflexes. Babinski sign.
                 Internuclear ophthalmoplegia.
    Impression: Probable multiple sclerosis.
    Plan: MRI of brain. 
          Start intravenous methlylprednisolone.
}
\end{Verbatim}

\textit{Note Standardization}: The GPT-4 API was iteratively prompted to standardize each source note with the following prompt:

\begin{Verbatim}[fontsize=\footnotesize]
prompt =
(You are a highly skilled medical terminologist 
specializing in clinical note 
standardization.  Your task is to standardize
this note and adhere to guidelines:

1. Expand Abbreviations:
   - Expand abbreviations to full words 
     (e.g., BP -> blood pressure).
   - For common abbreviations, expand 
     and retain  in parentheses 
     (e.g., MRI -> magnetic resonance 
      imaging (MRI)).
   - Resolve context-specific terms 
     (e.g., MS -> multiple sclerosis 
     (MS)).

2. Correct Spelling and Grammar:
   - Fix errors while preserving clinical 
     meaning.

3. Organize into Sections:
   - Use these headings: HISTORY, VITAL 
     SIGNS, EXAMINATION, LABS, RADIOLOGY, 
     IMPRESSION, and PLAN.
   - Move content to the correct sections
      and retain all data.

4. Standardize Terminology:
   - Replace non-standard terms with 
     standard terms (e.g., heart attack 
     -> myocardial infarction).

Standardize the provided clinical note as 
accurately as possible.
\end{Verbatim}

Output was saved as a JSON-compatible object with the following structure:
\begin{Verbatim}[fontsize=\footnotesize]
{
   "HISTORY": {
      "Chief Complaint": "String",
      "Interim History": "String"
   },
   "VITAL SIGNS": {
      "Blood Pressure": "String",
      "Pulse": "String",
      "Temperature": "String",
      "Weight": "String"
   },
   "EXAMINATION": {
      "Mental Status": "String",
      "Cranial Nerves": "String",
      "Motor": "String",
      "Sensory": "String",
      "Reflexes": "String",
      "Coordination": "String",
      "Gait and Station": "String"
   },
   "LABS": "String",
   "RADIOLOGY": "String",
   "IMPRESSION": {
      "Assessment": "String"
   },
   "PLAN": {
      "Testing": "String",
      "Education Provided": {
         "Instructions": "String",
         "Barriers to Learning": "String",
         "Content": "String",
         "Outcome": "String"
      },
      "Return Visit": "String"
   },
   "Metrics": {
      "Grammatical Errors": "integer",
      "Abbreviations Expanded": ["String"],
      "Spelling Errors": ["String"],
      "Non-Standard Terms": ["String"]
   }
}    
\end{Verbatim}

\textit{Standardized Note Evaluation}: Standardized notes were evaluated for length in characters, spelling errors corrected, grammatical errors corrected, abbreviations and acronyms expanded, medical terms substituted for non-standard terms, note reorganization into `canonical' sections. Source notes were compared to standardized notes to evaluate text organization, spelling and grammar, abbreviation expansions, terminology standardization, and completeness on a five-point Likert-like scale (1 = poor to 5 = excellent). A subset of 20 source notes were compared to standardized notes to identify any details lost in normalization.

\textit{Semi-Structured Searches}: Selected elements of the standardized notes were passed to the GPT-4 API to look for ''medications" in the PLAN section and to look for ''signs and symptoms" in the HISTORY, EXAMINATION, and IMPRESSION sections.

\section{Results and Discussion}
\begin{figure}
    \centering
    \includegraphics[width=0.95\linewidth]{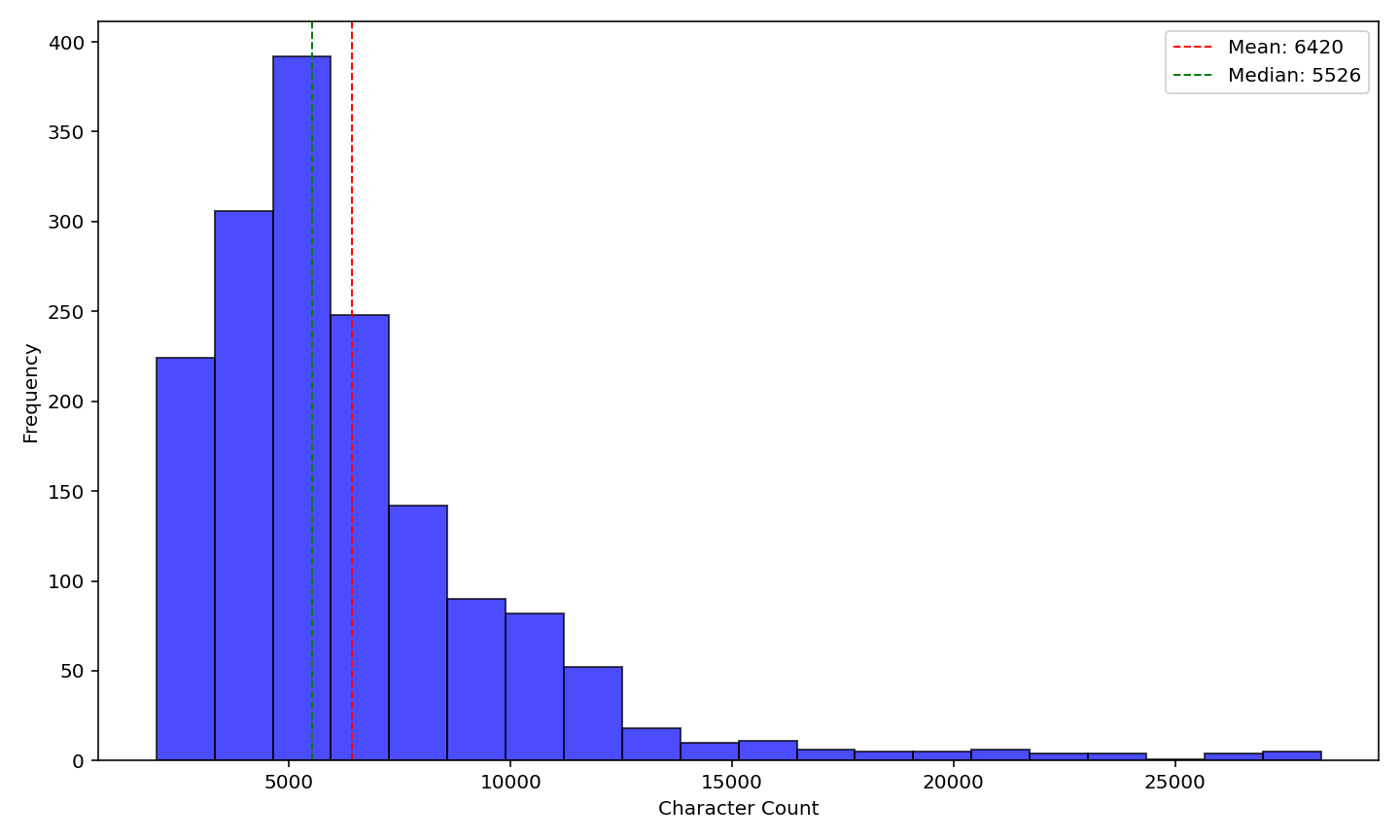}
    \caption{\textbf{Note Length}.~Mean note length in characters was 6420 $\pm$ 3691.}
    \label{fig:characters}
\end{figure}

We used a large language model (GPT-4) to standardize 1618 physician notes. GPT-4 was prompted to convert a source note into a standardized note. The source note was free text in ASCII format.  The standardized note was a formatted JSON object with canonical headings of HISTORY, EXAMINATION, IMPRESSION, and PLAN as well as appropriate sub-headings. GPT-4 was further prompted to correct spelling errors, expand abbreviations and acronyms, and disambiguate ambiguous abbreviations based on context. Mean note length was approximately 6,420 characters (Fig. \ref{fig:characters}). GPT-4 corrected 4.9 $\pm$ 1.8 grammatical errors per note, corrected 3.3 $\pm 5.2$ spelling errors, identified $3.1 \pm 3.0$ non-standard terms per note and made appropriate substitutions with medical terms, and expanded $15.8 \pm 9.1$ abbreviations and acronyms per note.\\

\begin{figure}
    \centering
    \includegraphics[width=0.95\linewidth]{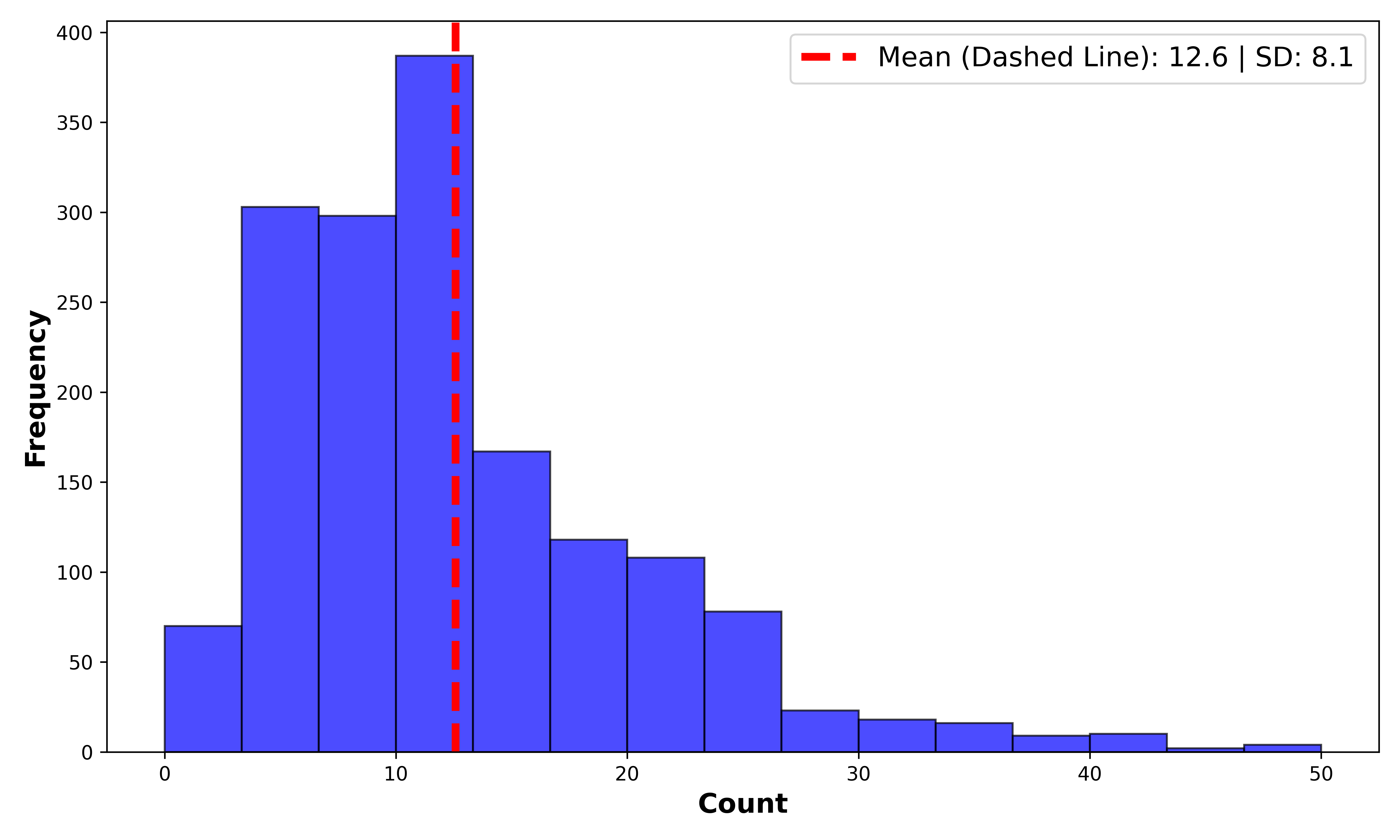}
    \caption{\textbf{Acronymns and Abbreviations}. ~Normalization expanded a mean of 15.8 $\pm$ 9.1 acronyms and abbreviations per note.}
    \label{fig:Abbreviations}
\end{figure}
All standardized notes were reviewed by a human expert for completeness, formatting, and accuracy. We used GPT-4 to compare the `source' notes to the standardized notes with regards to the following: text organization, spelling and grammar, abbreviation expansion, and terminology standardization. Measures were rated on a five-point Likert-like scale from 1 = `poor' to 5 = `excellent'. In a subset of 20 source notes that were reviewed in detail and compared to standardized notes, no loss of clinical content or detail was noted.

\begin{figure}
    \centering
    \includegraphics[width=0.95\linewidth]{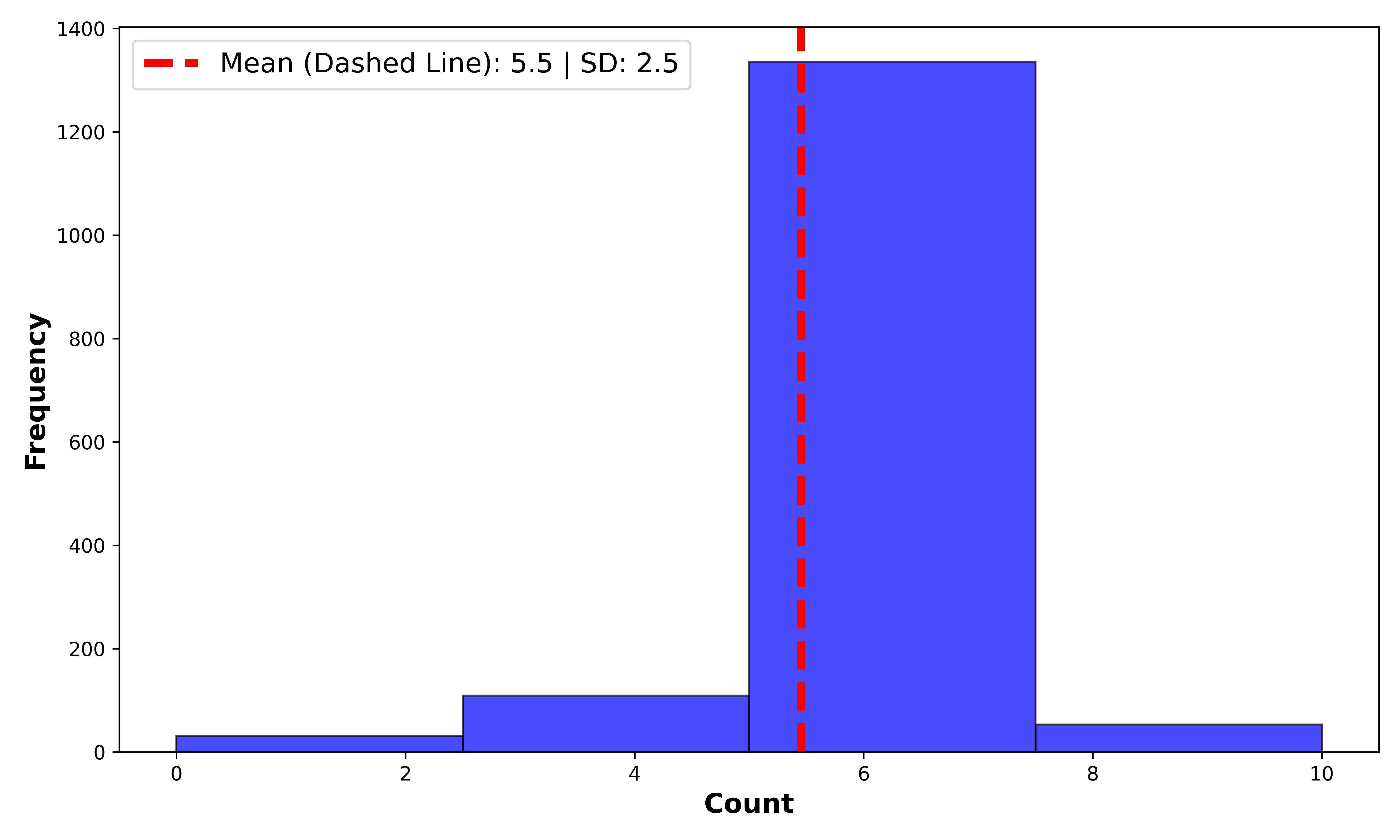}
    \caption{\textbf{Grammatical Errors.}~Note normalization corrected a mean of 4.9 $\pm$ 1.8 grammatical errors per note.}
    \label{fig:enter-label}
\end{figure}

\begin{figure}
    \centering
    \includegraphics[width=0.95\linewidth]{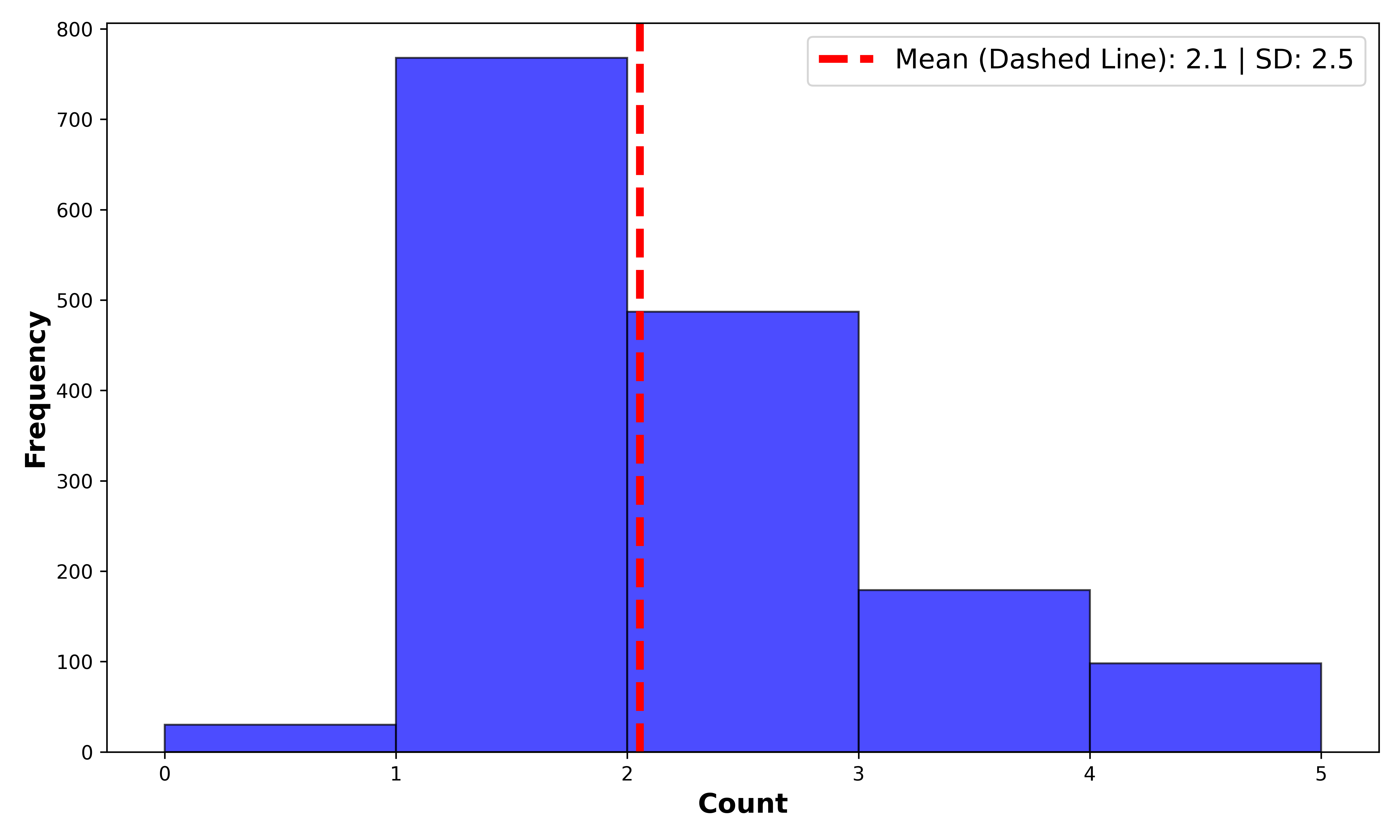}
    \caption{\textbf{Slang, Jargon, and Non-Standard Terms.}~~GPT-4 identified and corrected a mean of $3.1 \pm 3.0$ non-standard terms per note. An example of a non-standard term substitution is  ``feeling blue" $\rightarrow$ ''symptoms of depression" }
    \label{fig:non-standard terms}
\end{figure}

\begin{figure}
    \centering
    \includegraphics[width=0.95\linewidth]{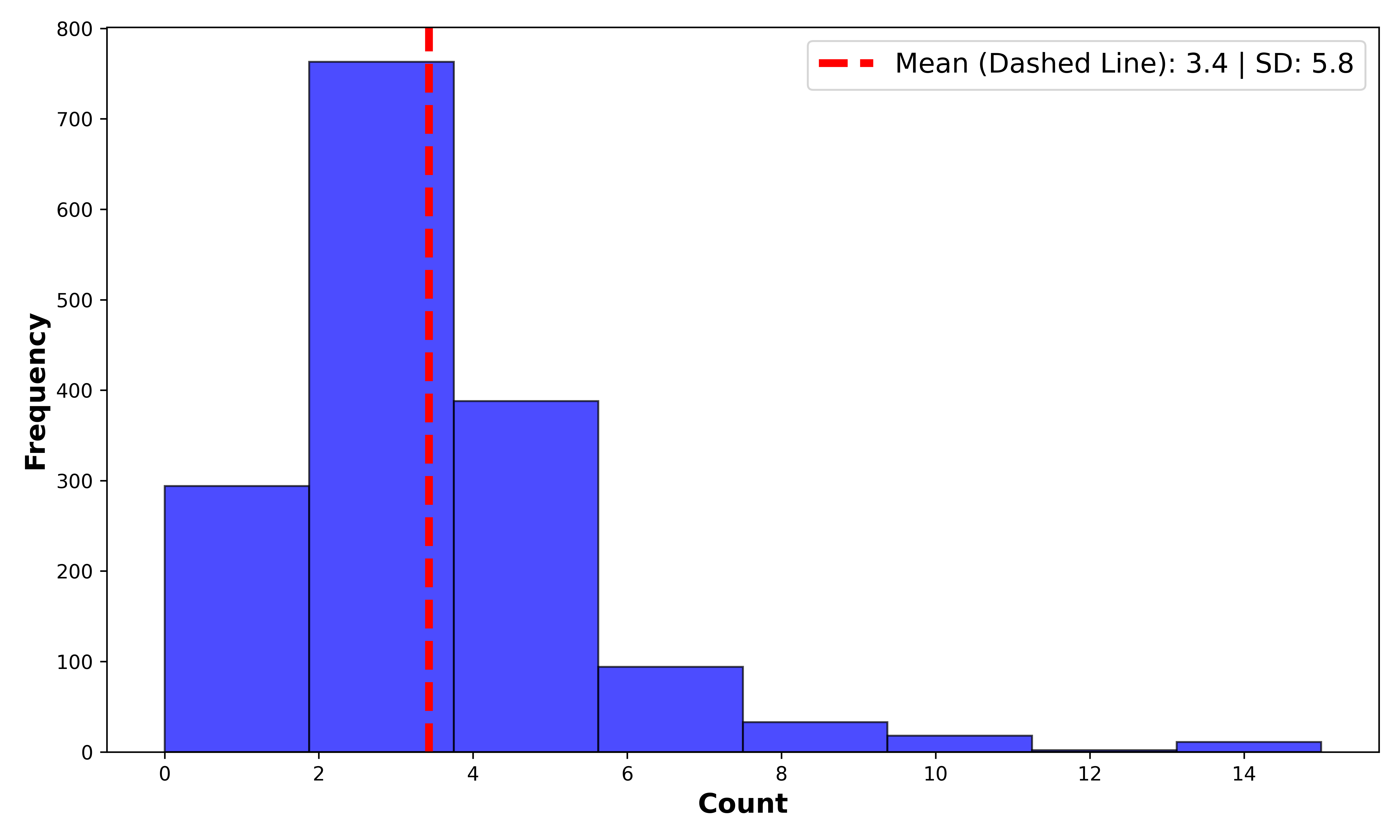}
    \caption{\textbf{Spelling Errors.} GPT-4 identified and corrected a mean of $3.3 \pm 5.2$ spelling errors per note. }
    \label{fig:spelling errors}
\end{figure}

\begin{figure}[ht]
    \centering
    \includegraphics[width=\linewidth]{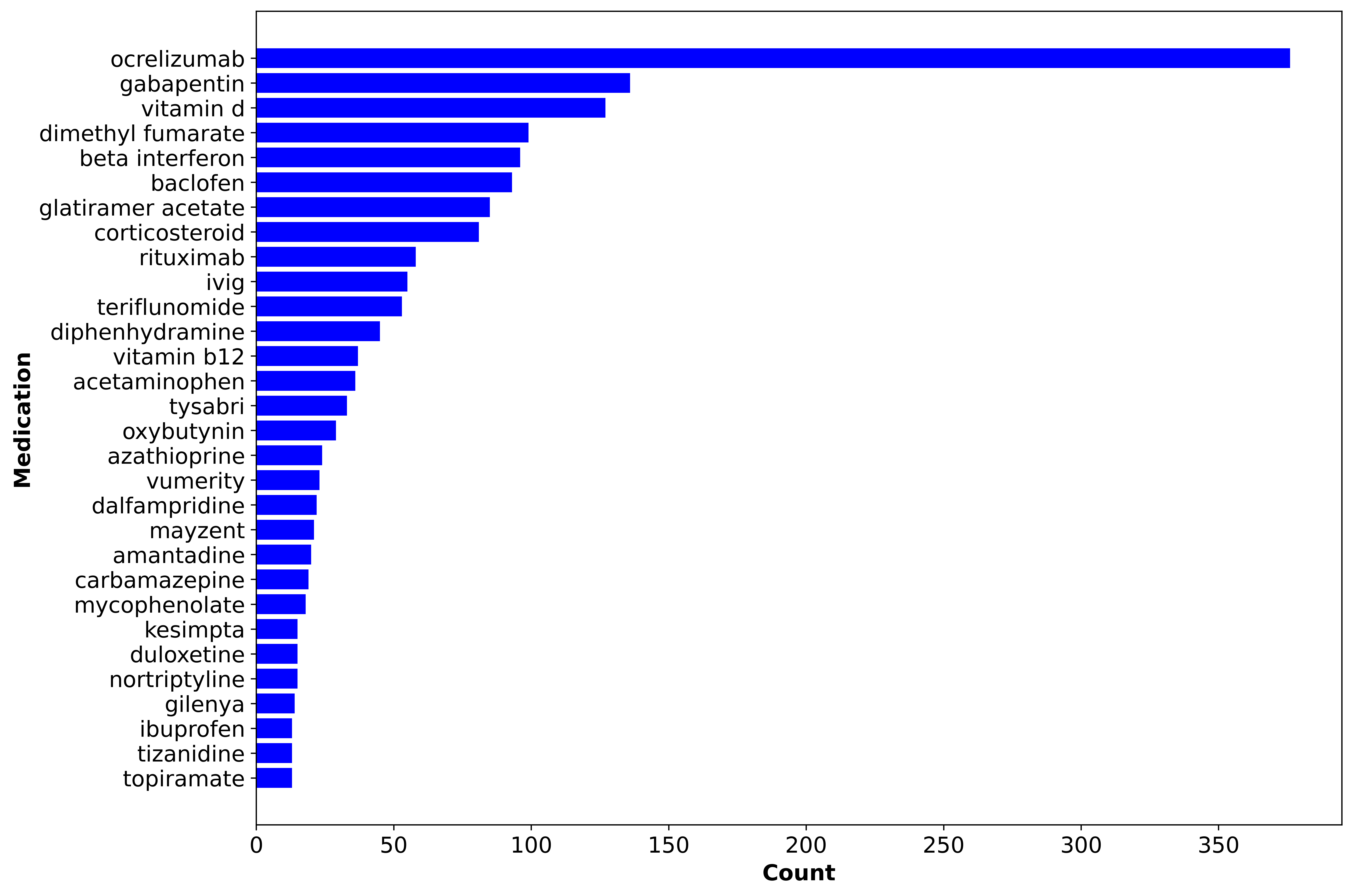}
    \caption{Medications found in the PLAN section of standarized notes. Most commonly prescribed medications include ocrelizumab (immunomodulator), gabapentin (pain management), dimethyl fumarate (immunomodulator), and baclofen (spasticity). These medications align with the observed prevalence of multiple sclerosis in the dataset.}
    \label{fig:Medications}

    \vspace{0.5cm} 

    \includegraphics[width=\linewidth]{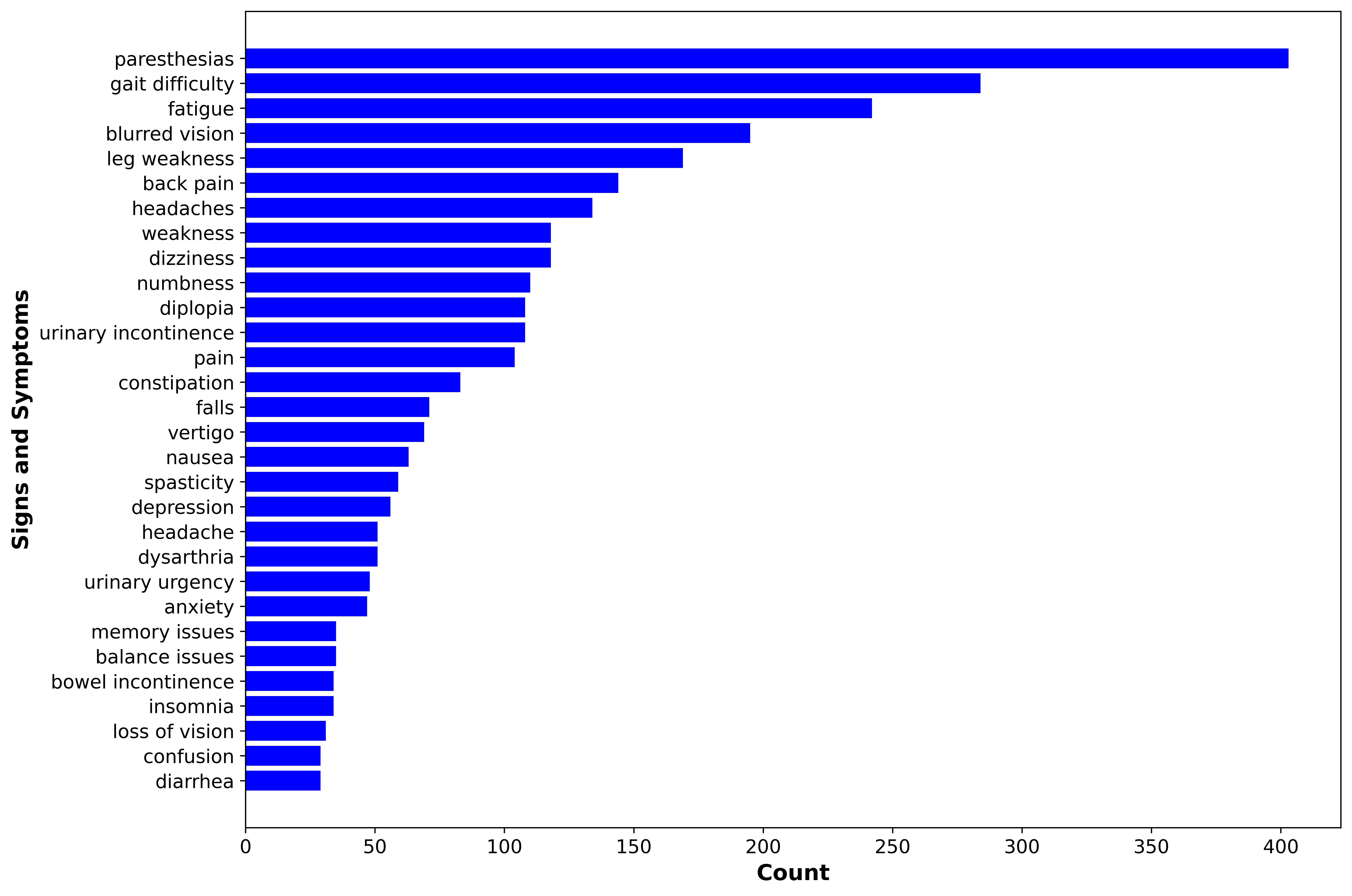}
    \caption{Signs and symptoms extracted from the HISTORY, EXAMINATION, and IMPRESSION sections of standarized notes. Commonly observed symptoms include paresthesias, gait difficulties, fatigue, blurred vision, and leg weakness—consistent with neurological diagnoses such as multiple sclerosis.}
    \label{fig:SignsSymptoms}
\end{figure}

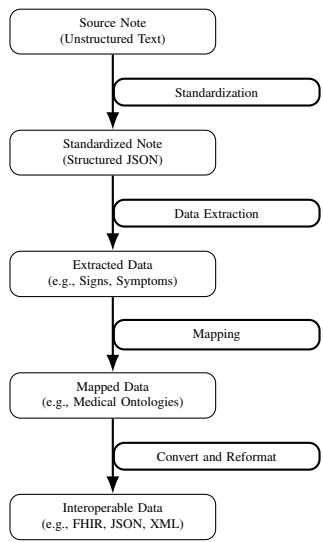
\begin{figure}[ht]
    \centering
    \begin{tikzpicture}[
        node distance=1cm and 1.5cm,
        every node/.style={draw, text width=2.5cm, align=center, font=\tiny, rounded corners},
        arrow/.style={-{Latex}, thick}
    ]
        \node (source) {Source Note\\(Unstructured Text)};
        \node (standarized) [below=of source] {Standardized Note\\(Structured JSON)};
        \node (extract) [below=of standarized] {Extracted Data\\(e.g., Signs, Symptoms)};
        \node (standardize) [below=of extract] {Mapped Data\\(e.g., Medical Ontologies)};
        \node (exchange) [below=of standardize] {Interoperable Data\\(e.g., FHIR, JSON, XML)};

        \draw [arrow] (source) -- (standarized) node[midway, right] {Standardization};
        \draw [arrow] (standarized) -- (extract) node[midway, right] {Data Extraction};
        \draw [arrow] (extract) -- (standardize) node[midway, right] {Mapping};
        \draw [arrow] (standardize) -- (exchange) node[midway, right] {Convert and Reformat};
    \end{tikzpicture}
    \caption{\textbf{Workflow for Structured Data Extraction, Mapping, and Reformatting.}~Steps from note standardization to creating interoperable data formats such as JSON and FHIR.}
    \label{fig:structured_workflow}
\end{figure}

By standardizing unstructured clinical notes into semi-structured standarized notes, we were able to use GPT-4 to perform semi-structured data retrieval on designated parts of the standarized note (Fig. \ref{fig:structured_workflow}. We were able to extract signs and symptoms from the HISTORY, EXAMINATION, and IMPRESSION sections of the notes (Fig. \ref{fig:SignsSymptoms}) and planned medications from the PLAN section of the standarized note (Fig. \ref{normalization_metrics}). Note normalization enables the extraction of clinically relevant signs, symptoms, and findings from unstructured text. These findings can be mapped to concepts in appropriate medical ontologies such as SNOMED CT or these findings can be converted into HL7 FHIR-compatible `observations,' facilitating seamless data exchange \cite{tayefi2021challenges, bender2013hl7, vorisek2022fast}. 

While large language models offer promising solutions to some challenges in the usability and quality of clinical notes, significant systemic issues remain beyond their current capabilities. One persistent problem is the widespread use of copy-and-paste practices by clinicians, which can propagate outdated, erroneous, or irrelevant content. Large language models have limited ability to discern whether copy-pasted content is clinically appropriate or needs to be updated \cite{siegler2009copy}. Similarly, large language models cannot reliably identify questionable documentation practices, such as omitting care rendered or documenting care that was not performed \cite{weiner2020accurate, sharma2008questionable}. Another challenge arises from the carry-forward of data, such as medication lists or problem histories, which may persist in notes without verification. Large language models cannot ascertain the accuracy of such information, particularly when clinical context is insufficient. While standardization improves note readability and usability, it does not inherently reduce documentation burden. Addressing this burden requires systemic changes, including improved user interface design, revised documentation guidelines, clinician training, and organizational support \cite{bakken2019can, kapoor2019physician, kang2023interventions, rodriguez2022s}. Although large language models can often disambiguate ambiguous terms and abbreviations based on context, they cannot resolve these ambiguities when the context is unclear or incomplete. These limitations highlight the need for complementary approaches to clinical documentation. Systemic improvements must address fundamental questions, such as workforce deployment (who documents what), documentation requirements (what needs to be documented), and workflows (how documentation occurs). These efforts, combined with technological advances, are essential to improving the quality and efficiency of clinical documentation.

This study has several limitations. The dataset was limited to 1618 physician notes from a neurology clinic, primarily with a diagnosis of multiple sclerosis. It is desirable to generalize these findings to a larger corpus of notes, with more diverse diagnoses,  with different note types, and from varied clinical settings.

As proof-of-concept, we demonstrated that standardized notes could be used for semi-structured data retrieval so that we were able to extract medications (Fig. \ref{fig:Medications}) and signs and symptoms (Fig. \ref{fig:SignsSymptoms}) from the notes. However, developing workflows that streamline the extraction of clinical information from standardized notes and converting key findings (e.g., signs, symptoms, diagnoses) into either concepts from medical ontologies (e.g., LOINC, SNOMED CT, RxNorm, etc.) or into HL7 FHIR-compatible resources will require significant additional effort and validation.

We did not perform a detailed analysis of the computational costs or processing times for note normalization. Preliminary estimates, based on consumer-grade workstations and the GPT-4 API, indicate an average processing time of approximately 20 seconds per note, with costs ranging from $0.01$ to $0.10$ USD per note, assuming typical note lengths of $500$ to $1,000$ tokens. These figures align with current GPT-4 pricing but may vary depending on note complexity, token usage, and future pricing models. While these estimates suggest that note normalization is feasible, scaling to high-volume clinical environments (e.g., $1,000$ to $10,000$ notes per day) will require further analysis and optimization to manage costs.

Our study used de-identified clinical notes, ensuring adherence to HIPAA privacy standards. However, this approach did not address the need to obtain explicit consent from physicians or patients. While de-identification mitigates many privacy concerns, future studies should implement additional safeguards to address patient privacy. Note identification is not possible in a active clinical setting. Furthermore, when physician notes are reconfigured, explicit consent or institutional approval from clinicians may be necessary.

Note standardization transforms source notes, often stored as unstructured ASCII text, into structured formats with canonical headings. This demonstrates the potential of reconfiguring clinical notes to meet diverse needs. The field of clinical documentation is increasingly focused on note templates, template engineering, and note quality improvement \cite{epstein2021effect, savoy2021clinical, ebbers2022impact, hochmankatherine2024scaling, burke2014qnote}. Once clinical text is digitized, large language models can reformat notes into various styles and templates tailored to specific use cases, enhancing their adaptability and utility.

\section{Conclusions}
Note standardization using large language models offers a promising approach that enhances the readability, usability, quality, and interoperability of physician notes. Large language models excel at improving note structure, expanding abbreviations, replacing slang and jargon with standardized terminology, and correcting spelling and grammatical errors while preserving clinical content and clinician intent. By standardizing clinical notes, large language models can facilitate semi-structured data retrieval, ready notes for the mapping of terms to concepts in medical ontologies, and prepare notes for compliance with emerging interoperability standards such as FHIR \cite{vorisek2022fast}.
\bibliographystyle{IEEEtran}
\bibliography{references}
\end{document}